\documentclass{IEEEoj}
\usepackage{cite}
\usepackage{amsmath,amssymb,amsfonts}
\usepackage{algorithm}
\usepackage{algpseudocode}
\usepackage{amsmath}

\usepackage{graphicx,color}
\usepackage{textcomp}
\usepackage{changepage}
\usepackage{url}
\usepackage{graphicx}
\usepackage{subcaption}
\usepackage{hyperref}
\usepackage{hyperref}
\usepackage{comment}
\usepackage{graphicx}
\usepackage{caption}
\usepackage{subcaption}
\usepackage{changepage}
\usepackage{cite}
\usepackage{amsmath,amssymb,amsfonts}

\usepackage{graphicx}
\usepackage{textcomp}
\usepackage{subcaption}
\usepackage{booktabs} 
\usepackage{multirow}
\usepackage{hyperref}

\def\BibTeX{{\rm B\kern-.05em{\sc i\kern-.025em b}\kern-.08em
   T\kern-.1667em\lower.7ex\hbox{E}\kern-.125emX}}
\AtBeginDocument{\definecolor{ojcolor}{cmyk}{0.93,0.59,0.15,0.02}}
\def\OJlogo{\vspace{-14pt}\includegraphics[height=28pt]{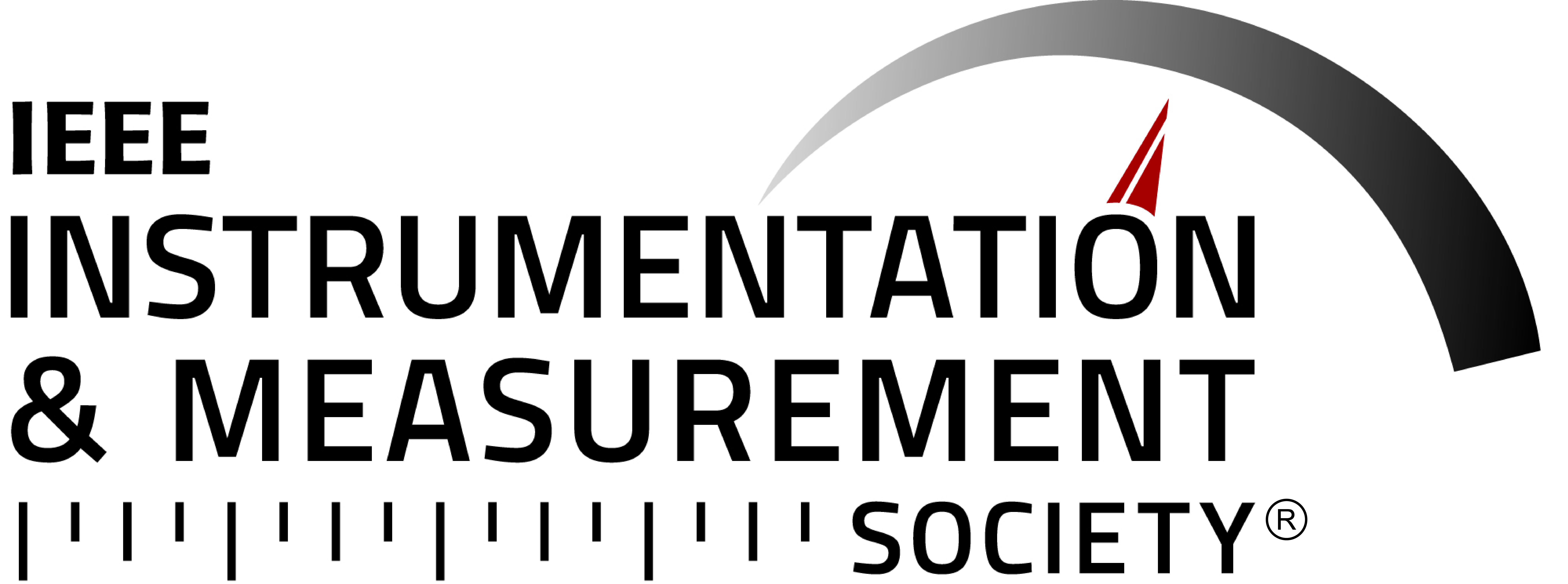}}
\begin{document}
\receiveddate{XX Month, XXXX}
\reviseddate{XX Month, XXXX}
\accepteddate{XX Month, XXXX}
\publisheddate{XX Month, XXXX}
\currentdate{XX Month, XXXX}
\doiinfo{OJIM.2022.1234567}
\title{Brain Tumor Classification from MRI Scans via Transfer Learning and Enhanced Feature Representation}
\author{
\uppercase{Ahta-Shamul Hoque Emran\authorrefmark{1}}, 
\uppercase{Hafija Akter\authorrefmark{2}}, 
\uppercase{Abdullah Al Shiam\authorrefmark{3}}, 
\uppercase{Abu Saleh Musa Miah\authorrefmark{4}},
\uppercase{Anichur Rahman\authorrefmark{5}}, 
\uppercase{Fahmid Al Farid\authorrefmark{5,6}}, 
\uppercase{Hezerul Abdul Karim\authorrefmark{6}}, 
\IEEEmembership{Member, IEEE}
}
\affil{\authorrefmark{1}Department of CSE, Netrokona University, Netrokona, Bangladesh (e-mail: 201904002@neu.ac.bd)}
\affil{\authorrefmark{2}Department of CSE, Netrokona University, Netrokona, Bangladesh (e-mail: 201904023@neu.ac.bd)}
\affil{\authorrefmark{3}Department of CSE, Netrokona University, Netrokona, Bangladesh (e-mail: shiam.cse@neu.ac.bd)}
\affil{\authorrefmark{4}Department of CSE, The University of Rajshahi, Rajshahi, Bangladesh (e-mail: musa.cse@ru.ac.bd)}
\affil{\authorrefmark{5}Department of Information Technology, Allen E. Paulson College of Engineering and Computing, Georgia Southern University, Statesboro, GA 30458, Georgia, USA (e-mail: ar36248@georgiasouthern.edu )}
\affil{\authorrefmark{6}Centre for Image and Vision Computing (CIVC), COE for Artificial Intelligence, Faculty of Artificial Intelligence and Engineering (FAIE),  Multimedia University, Cyberjaya 63100, Malaysia (e-mails: fahmid.alfarid@mmu.edu.my, hezerul@mmu.edu.my )}

\corresp{Corresponding authors:Hezerul Abdul Karim (hezerul@mmu.edu.my ),Fahmid Al Farid (fahmid.alfarid@mmu.edu.my), Abdullah Al Shiam (shiam.cse@neu.ac.bd).}
\authornote{This work was supported by Multimedia University, Cyberjaya, Selangor, Malaysia (Grant Number: PostDoc(MMUI/240029)).}
\markboth{Preparation of Papers for IEEE OPEN JOURNALS}{Author \textit{et al.}}
\begin{abstract}
Brain tumors are abnormal cell growths in the central nervous system (CNS), and their timely detection is critical for improving patient outcomes. This paper proposes an automatic and efficient deep-learning framework for brain tumor detection from magnetic resonance imaging (MRI) scans. The framework employs a pre-trained ResNet50 model for feature extraction, followed by Global Average Pooling (GAP) and linear projection to obtain compact, high-level image representations. These features are then processed by a novel Dense-Dropout sequence, a core contribution of this work, which enhances non-linear feature learning, reduces overfitting, and improves robustness through diverse feature transformations.
Another major contribution is the creation of the Mymensingh Medical College Brain Tumor (MMCBT) dataset, designed to address the lack of reliable brain tumor MRI resources. The dataset comprises MRI scans from 209 subjects (ages 9–65), including 3,671 tumor and 13,273 non-tumor images, all clinically verified under expert supervision. To overcome class imbalance, the tumor class was augmented, resulting in a balanced dataset well-suited for deep learning research. This curated resource provides a strong foundation for advancing brain tumor diagnosis, classification, and detection. Evaluated on both the newly created MMCBT dataset and the publicly available BTM benchmark dataset, the proposed framework demonstrates exceptional reliability. It achieves accuracies of 97.57\% on the MMCBT dataset and 99.80\% on the BTM dataset, along with consistently strong precision, recall, and F1-scores—metrics crucial for minimizing diagnostic errors in medical imaging. These results highlight the novelty of the framework and its potential as a robust clinical decision-support tool.
\end{abstract}


\maketitle

\section{Introduction}
\label{sec:introduction}
\PARstart{A}
 brain tumor is a collection of abnormal cells in the brain\cite{ref41}. Brain tumors stand  as a major concern in the field of medicine owing to the well-known complications and  dire consequences. An early diagnosis is essential for successful treatment and long-term  survival in patients with a brain tumor \cite{ref18}. Historically, the diagnosis of brain tumors  required human interpretation of medical images, and this process was not only slow but also susceptible to human errors. Recent advancements in technology have changed the  game in brain tumor detection, and now, thanks to tools like Magnetic Resonance Imaging  (MRI) and Convolutional Neural Networks (CNNs), it's never been easier to identify and diagnose brain tumors. The rapid development of deep learning technologies has greatly influenced the field of medical imaging, particularly with regard to detecting complex diseases such as brain tumors. The opportunity to use very powerful computational models in the analysis of medical image scans, in particular MRI scans, and the way they will transform diagnostic procedures. Early detection is crucial for increasing the survival rate of patients with brain tumors \cite{ref18}; therefore, automatic and accurate brain tumor detection is essential for a better healthcare system \cite{ref43}\cite{ref45}\cite{ref46}. Deep learning-based techniques have shown promising results for the detection and classification of brain tumors with high accuracy, thereby minimizing dependence on human expertise and time to diagnosis. Brain tumor detection relies on neural activities originating from thecomplexity and diversity of the human brain. These signals of the brain are interpreted through medical images such as MRI and CT Scan. These
image modalities represent the spatial and intensity data of the brain tissues that appear in and around the tumor. The brain contains
different types of tissues (gray matter, white matter, cerebrospinalfluid), and tumors may be benign or malignant \cite{ref7}\cite{ref9}. Neural activities that contribute to brain tumor differentiation rely on unusual tissue consolidation arrangements, texture distortions, shape, and abnormal ink shades in brain images \cite{hassan2024residual_miah_alzh,hassan2025neurological,hassan2025stacked_alzh_miah,Hassan_alzh_gradual_miah_,haque2024multi_heart_disease}. The neural activity of the brain contributes significantly to the process of brain tumor formation, as well as to the detection through neuroimaging. Brain cells, the neurons, communicate with one another via electrical signals that govern basic body processes such as movement,
memory, and pain sensation \cite{ref38}. Erratic neural behavior, often resulting from brain tumors, may present in the form of seizures, paralysis, cognitive loss, and other neurologic signs \cite{miah2025methodologica_pd,shin2025autism_miah,shin2025_pd_musa,matsumoto2025machine_musa_pd}.  In Imaging techniques, when MRI used to locate a brain tumor, it reveals physical changes to the structure of the brain such as tissue abnormalities, swelling or mass effect which is the
pressure of a tumor on the surrounding brain. Furthermore, we can use MRI to measure changes in brain activity based on blood flow to specific locations in the brain \cite{ref35}, that helps determine how a tumor is affecting neural
activities and interacting with other functions of the brain.  This study aims to propose a deep learning-based method for brain tumor detection. By  introducing Global Average Pooling (GAP) based feature extraction and Linear Projection  (LP) based dimensionality reduction, the study seeks to provide improved overall accuracy,  efficiency, and interpretability of brain tumor detection. The adoption of GAP allows for  pooling of the global image features, and therefore, the computational cost is reduced
 compared with traditional methods.
 
 Despite significant progress in brain tumor detection, several challenges remain in achieving high-accuracy and high-efficiency diagnoses \cite{ref9,ref38,ref35}. MRI data is highly dimensional, making it computationally expensive to process. Additionally, datasets may have an imbalanced class distribution, leading to model bias, and many deep learning models are considered "black boxes," making it difficult for clinicians to trust their outputs. 
The main contributions of this work are summarized as follows:  
\begin{itemize}  
    \item \textbf{Proposed Framework:} We introduce an automatic and efficient deep-learning framework for brain tumor detection from MRI scans, leveraging ResNet50 with Global Average Pooling (GAP) and linear projection for feature extraction, along with a novel Dense-Dropout sequence to enhance non-linear learning, reduce overfitting, and improve model robustness.  

    \item \textbf{New Dataset:} We created the \textit{Mymensingh Medical College Brain Tumor (MMCBT) dataset}, consisting of MRI scans from 209 subjects (ages 9--65), including 3,671 tumor and 13,273 non-tumor images, clinically verified under expert supervision. The dataset was balanced through augmentation and provides a valuable resource for advancing research in brain tumor analysis.  

    \item \textbf{Performance Achievement:} The proposed framework achieves state-of-the-art performance, with accuracies of \textbf{97.57\%} on the MMCBT dataset and \textbf{99.80\%} on the public BTM dataset, along with strong precision, recall, and F1-scores, underscoring its potential as a robust tool for clinical decision support.  
\end{itemize}


\section{Literature Review}
Recent studies have focused on using deep learning, especially convolutional neural networks (CNN), for the detection of brain tumors from magnetic resonance images.These methods are proven to be accurate and efficient with minimal reliance on human expertise. But some challenges, such as high-dimensional data, class imbalances, and model interpretability, still limit their complete clinical application. Early diagnosis is still a key factor in treatment outcome and patient survival.
Ghazanfar Latif et al. \cite{ref41} proposed a methodology to classify brain
glioma tumor using Block-Based 3D Wavelet Features of MR Images. They
have accomplished this work by multiclass. In this research study, the
proposed Random Forest classifier, HGG and LGG respectively. In this
work, they used 18600 MR images. Also, they acquired
89.75\%,86.87\%accuracy by using Random Forest classifier, HGG and LGG
respectively that accuracy is very low. Neelum Noreen et al. \cite{ref42} proposed a methodology to diagnosis brain
tumor a deep Learning model based on concatenation. They have
accomplished this work by Random features extraction. In this research
study, the proposed Inception-v3 and DensNet201 classifier. This work
used brain tumor dataset Figshare, which is publicly available. In this
work, they produced 99.34 \%, and 99.51\% testing accuracies
respectively with Inception-v3 and DensNet201. As they acquired high
accuracy but they did not mention their Newly Created MMCBT Dataset. Arashdeep Kaur \cite{ref43} proposed a methodology to extract brain tumor by
using different segmentation method. He used a dataset of 25 MR Images
of brain that used for testing the system and recording the experimental
results. He did not mention any clear feature extraction method but he
used Fuzzy c-means algorithm and acquired 90.57\% accuracy that is very
low accuracy. Gayathri S et al. \cite{ref44} proposed a methodology to analyze, detect,
and automatic classification of different stages of brain tumor. In this
work, they used Support Vector Machine (SVM) classification method. They
extract features such as color, shape and texture. In this work they use
multiple data but did not mention any clear dataset. They acquired
98.03\% accuracy. Khizar Abbas et al. \cite{ref45} proposed a methodology to detect brain
tumor using machine learning. They proposed LIPC based methodology for
brain tumor classification and segmentation. They extract features such
as vector size, standard deviation, histogram, kurtosis, and skewness.
In this work, they used 3D brain tumor dataset. In this research work,
they acquired 95\% accuracy by using LIPC but they did not mention how
much dataset they used for their research.

Sahar Ghanavati et al. \cite{ref46} proposed a methodology to detect brain
tumor by using MRI. In this research study, they use a multi-modality
framework for automatic tumor detection. They did not mention any clear
idea of dataset. They extract features such as intensity, shape
deformation, symmetry, and texture features. In this work, they used
Automatic tumor detection algorithm and acquired average accuracy of
about 90\% and that was not good. K N Deeksha et al. \cite{ref47} proposed a methodology to classify brain
tumor. In this work, they extract feature such as color, size, location,
edge, shape. They used dataset which was collected from the website
figshare, and it consists of 2,123 images of MRI scans of the brain
although it's not their Newly Created MMCBT Dataset. In this research study, they
used CNN algorithm and they acquired 93\% for the training dataset and
92\% for the testing dataset. Mahdis Roshanfekr Rad et al. \cite{ref48} proposed a methodology to detect
brain tumor. In this work, they used 3D MRI dataset but how much dataset
they used it's not mentioned. They extract features such as lines,
edges, and subjective contours. In this research study, they used (GLCU)
and artificial neural network (ANN) Classification method. They acquired
98\% accuracy that's very high. ADEKANMI A. ADEGUN et al. \cite{ref49} proposed a methodology to detect and
classify of skin lesions in dermoscopy images. The proposed model was
evaluated on publicly available HAM10000 dataset of over 10000 images
consisting of 7 different categories. In this work, they extract feature
such as Local Binary Pattern (LBP), Edge Histogram (EH), Histogram of
Oriented Gradients (HOG) and Gabor methods to extract color, texture,
and shape. In this research study, they used FCN-based DenseNet
classification method. They acquired 98\% accuracy that's very high. V.P. GLADIS PUSHPA RATHI et al.\cite{ref50} proposed a methodology to
detect and characterized of brain tumor. In this work, they used HSOM,
Wavelet packet feature extraction method. They used MRI dataset but how
much dataset they used it's not mentioned. In this research study, they
used ANN classification method and they required 96.6\% accuracy. Khaleda Akhter Sathi et al.\cite{ref51} proposed a methodology to classify
brain tumor. In this work, they used Gabor + DWT + GLCM feature
extraction method. They used a total of 180 images database for
classification which is not huge. In this research study, they used ANN
classification method and they required 98.9\% accuracy which is very
high. Sanjay Kumar C et al. \cite{ref52} proposed a methodology to categorized
brain tumor. In this work, they used 26 features out of which 13 are
local and 13 are global are computed from GLCM and are utilized in
training. They did not mention any clear idea about dataset. In this
research study, they used SVM with Hybridized Local-Global Features
classification method and they required 95\% accuracy.

Muhammad Nazir et al. \cite{ref53} proposed a methodology to detect brain
tumor. In this work, they used Haar wavelet, DWT and DCT feature
extraction method. They did not mention any clear idea about dataset. In
this research study, they used DCT+NN and DCT+SVM classification method
and they required 94.5\% and 95\% accuracy respectively. Hein Tun Zaw et al. \cite{ref54} proposed a methodology to detect brain
tumor. A total of 114 MRI images with 24 normal and 90 tumors are used
for this research work. They extract feature such as used thresholding,
edge detection, histogram equalization, segmentation. In this work, they
used Naïve Bayes classification method and they acquired 94\% accuracy. M. Monica Subashini et al. \cite{ref55} proposed a methodology to analyze
brain tumor. In this research work they used MRI images but did not
mention any clear idea about data set how much dataset they used for
their research. Also, they did not mention any clear idea about
classification method and accuracy which is not good. Kang Han Oh et al. \cite{ref56} proposed a methodology to detect brain tumors. 
\cite{saeedi2023mri} presents two deep learning methods and several machine learning approaches for accurately diagnosing brain tumors using MRI images. The proposed 2D Convolutional Neural Network achieved a training accuracy of 96.47\%, demonstrating optimal performance compared to other methods, making it a valuable tool for clinical use
In this research work, the scheme successfully detects the tumor
region on the 60-brain magnetic resonance dataset. \cite{asiri2024optimized} introduces a novel two-module computerized method for brain tumor detection and classification from MRI images. The first module enhances images using adaptive Wiener filtering, neural networks, and independent component analysis, while the second module employs Support Vector Machines (SVM) for segmentation and classification. This research \cite{vasavi2025hybrid} addresses the challenging problem of brain tumor detection from MRI images by developing a novel deep learning architecture called EfficientNet-DbneAlexnet that combines EfficientNet with Deep batch normalized eLUAlexnet.
In this work, they
used SVM classification method but they did not mention any clear idea
about accuracy which was very needed.

\section{Dataset}
In the study we used two dataset to evaluate the proposed model which describe below. 
\subsection{Newly Created Mymensingh Medical College Brain Tumor (MMCBT) Dataset}
This dataset consists of brain tumor MRI images, collected from Mymensingh Medical College, Mymensingh, Bangladesh. The images were initially obtained in DICOM format and subsequently converted into JPEG format using Onis Viewer software for further analysis. The subjects ranged in age from 9 to 65 years, and included both male and female participants. Table \ref{tab:MMCBT_data_table} shows the summery of the newly created MMCBT dataset. In total, 209 subjects were considered, among which 63 were diagnosed with brain tumors and 146 were identified as non-tumor cases.  The dataset contains 3,671 images of tumor cases and 13,273 images of non-tumor cases, each acquired at different imaging angles. All images were carefully labeled and verified by Dr. Md. Kamal Hossen, MBBS, BCS (Health), Resident Surgeon, Department of Neurosurgery, Mymensingh Medical College Hospital, under clinical supervision to ensure high-quality and reliable annotations. To address the class imbalance, the tumor class was augmented to 13,252 images, making the dataset balanced for deep learning applications.
This curated dataset is particularly suitable for medical image analysis and deep learning studies, supporting research on the early diagnosis, classification, and detection of brain tumors.

\begin{table}[]
\caption{Summary of the Newly Created MMCBT MRI Dataset}
\label{tab:MMCBT_data_table}
\resizebox{\columnwidth}{!}{%
\begin{tabular}{|c|c|c|c|}
\hline
\textbf{Class} & \textbf{No. of Subjects} & \textbf{No. of Images} & \textbf{After Augmentation} \\ \hline
Tumor          & 63                       & 3,671                  & 13,252                      \\ \hline
Non-Tumor      & 146                      & 13,273                 & 13,273                      \\ \hline
\textbf{Total} & \textbf{209}             & \textbf{16,944}        & \textbf{26,525}             \\ \hline
\end{tabular}%
}
\end{table}

\begin{figure}[htbp]
    \centering
    \includegraphics[width=0.425\textwidth]{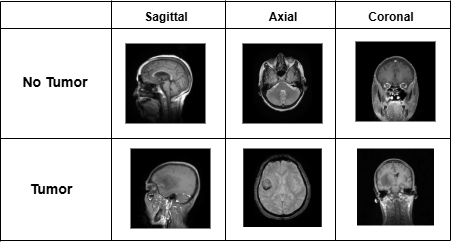}
    \caption{\centering Different Sample of Brain Tumor of Newly Created MMCBT dataset}
    \label{Figure of Dataset-I}
\end{figure}

\subsection{Brain Tumor Mendeley  (BTM) Dataset}
This dataset consists of Brain Tumor MRI images, collected from Mendeley
Data \cite{ref58}. In this dataset, original 7023 MRI images contained both
tumor and non-tumor class, where 5023 images of Tumor cases and 2000
images of non-tumor cases. To balance the dataset, we augmented the
non-tumor class data. Then, We have prepared our dataset, where non-tumor
class contains 5024 images and total 10047 images contain augmented dataset.

\begin{figure}[htbp]
    \centering
    \includegraphics[width=0.425\textwidth]{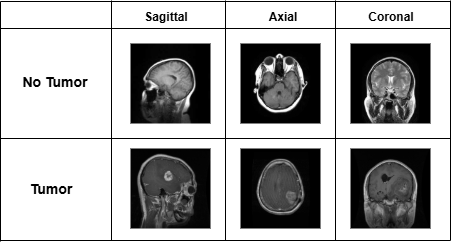}
    \caption{\centering Different Sample of Brain Tumor of public BTM Dataset}
    \label{figure of Dataset-II}
\end{figure}

\section{Proposed Method}
The journey of our proposed method begins with MRI scans, where brain images are first collected and prepared for analysis. To make sense of these complex images, we employ a pre-trained ResNet50 model, which acts as a skilled observer, carefully extracting meaningful patterns from the scans. Instead of handling raw features directly, these patterns are refined using Global Average Pooling (GAP) and linear projection, producing compact, high-level representations that capture the essence of the data.
Next, these representations are passed through a novel Dense-Dropout sequence, designed to act like a careful filter—strengthening useful signals, suppressing noise, and preventing the model from memorizing rather than learning. This stage ensures that the system generalizes well, even when faced with new and unseen cases. At the heart of this story lies the MMCBT dataset, a newly created collection of clinically verified MRI scans from 209 subjects. Balanced and comprehensive, it gives the model a strong foundation to learn the difference between tumor and non-tumor cases. Finally, when tested on both the MMCBT dataset and the public BTM benchmark, the method demonstrates outstanding performance.
\subsection{Pre-processing}
Preprocessing is an essential step in feature extraction from MRI images and begins after the image acquisition stage.
The collected brain MRI data were initially stored in DICOM (Digital Imaging and Communications in Medicine) format, which is the standard format used in medical imaging. To ensure compatibility with commonly used image processing frameworks, the DICOM files were converted into JPEG format using Onis Viewer.

For each subject, MRI scans were acquired in three standard anatomical planes: Axial (top-to-bottom view), Coronal (front-to-back view), Sagittal (left-to-right view).

Each plane consisted of a series of slice images, capturing different cross-sectional views of the brain. On average, approximately 27 slices per plane were collected, resulting in  81 images per subject. After conversion, the images were systematically organized and labeled according to subject ID and anatomical plane. This preprocessing step ensured data uniformity, reduced format-related complexity, and prepared the dataset for subsequent analysis and model training.
Then resized images to a fixed size for compatibility with neural network architectures, normalized pixel intensity values to standard ranges (e.g., $[0,1]$ or $[-1,1]$) to improve model convergence, and applied histogram equalization to enhance image contrast and highlight important features \cite{ref49,ref55}. Mathematically, the preprocessed image $\mathbf{P}$ can be expressed as $\mathbf{P} = h(\mathbf{I})$, where $h(\cdot)$ represents the set of preprocessing functions.

\subsection{Feature Extraction via Transfer Learning}
Feature extraction is an important step in which the system extracts the
relevant features from the MRI image. The proposed method uses Global
Average Pooling (\textbf{GAP}) and Linear Projection to densely subset
features.
\begin{figure*}[t!]
    \centering
    \includegraphics[width=\textwidth]{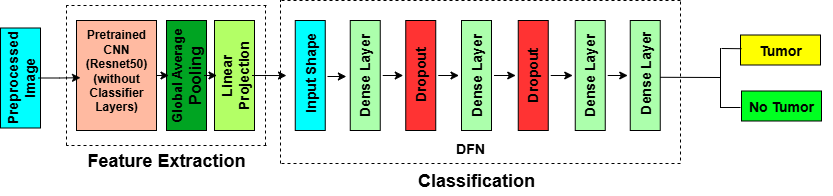}
    \caption{\centering Illustrate the Proposed Method for GAP-Based Feature
Extraction and Linear Projection.}
    \label{fig3}
\end{figure*}

To capture discriminative spatial features, a pre-trained ResNet50 backbone (initialized on ImageNet) was employed. Unlike training from scratch, transfer learning leverages pre-learned low-level filters while fine-tuning high-level representations for brain MRI data.
The final convolutional feature maps $F \in \mathbb{R}^{7 \times 7 \times 2048}$
were reduced via Global Average Pooling (GAP) to a 2048-dimensional vector:
\begin{quote}
\begin{equation}
\mathbf{F}_{\mathrm{GAP}} =
\frac{\mathbf{1}}{\mathbf{H} \times \mathbf{W}}
\sum_{i=1}^{\mathbf{H}} \sum_{j=1}^{\mathbf{W}} \mathbf{F}_{i,j}
\end{equation}
\end{quote}

where H=W=7. GAP offers interpretability, computational efficiency, and reduced risk of overfitting compared with flattening approaches.

\subsection{Linear Projection for Dimensionality Reduction}
The pooled features were projected into a compact 512-dimensional space using a dense layer:
\begin{quote}
\begin{equation}
\emph{\textbf{F\textsubscript{proj} = W . F\textsubscript{GAP}+ b}}
\end{equation}
\end{quote}
where \( W \in \mathbb{R}^{512 \times 2048} \) and b is the bias term. This projection reduces redundancy while retaining task-relevant information, ensuring scalability for large MRI datasets.

\begin{algorithm}[H]
\caption{Feature Extraction using ResNet50}
\label{pseudocode1}
{\textbf{Input:}- Preprocessed image $P$ of shape $(224,224,3)$\\
{\textbf{Output}:} 512-dimensional feature vector $F_{extracted}$\\}

Step 1: Load pre-trained ResNet50 model without top layer\\
$~~~~~$$\textbf{resnet50\_model} \gets \texttt{ResNet50(weights='imagenet', include\_top=False)}$\\

Step 2: Specify input tensor\\
$~~~~~$$\textbf{X} \gets \texttt{Input(shape=(224,224,3))}$\\

Step 3: Pass input through ResNet50 to get feature map\\
$~~~~~$$\textbf{F} \gets \texttt{resnet50\_model(X)}$\\

Step 4: Apply Global Average Pooling (GAP) on $\mathbf{F} \in \mathbb{R}^{7 \times 7 \times 2048}$\\
$~~~~~$$\mathbf{F_{GAP}} \gets \frac{1}{7 \times 7} \sum_{i=1}^{7} \sum_{j=1}^{7} F_{i,j}$ \\

Step 5: Apply a Dense layer to reduce dimension to 512\\
$~~~~~$$\textbf{F\_512} \gets \texttt{Dense(512)(F\_GAP)}$\\

Step 6: Define the final model\\
$~~~~~$$\textbf{Final\_Model} \gets \texttt{Model(inputs=X, outputs=F\_512)}$\\

Step 7: Extract features from input image $P$\\
$~~~~~$$\textbf{F\_extracted} \gets \texttt{Final\_Model(P)}$\\

Step 8: Return $F_{extracted}$\\
$~~~~~$$\textbf{Return} \, F_{extracted}$\\
\end{algorithm}

\subsection{Classification}

The extracted features are processed using a deep learning architecture, namely a deep feedforward network (DFN). While the primary experiments in this study are conducted with the DFN, additional experiments are also performed using CNN-1D and FCNN to further evaluate the robustness of the proposed approach.

\subsubsection{DFN}

Feedforward neural networks such as Multilayer Perceptron (MLP) are
artificial models where information flow strictly progresses from the
input through the hidden layers to the output. It is trained to estimate
a rich representation of the data, with each layer being fully connected
to the subsequent one. DFN can be applied to tasks such as
classification and regression by learning the mapping from input
features to output predictions.

In mathematical terms, each layer's output is obtained as a linear
combination of its input, which is then passed through an activation
function to provide the required non-linearity.
In a neural network, the pre-activation value of a layer is computed as:
\[
z^{(l)} = W^{(l)}a^{(l-1)} + b^{(l)}
\]
The activation function is then applied element-wise to generate the
layer's output:
\[
a^{(l)} = f(z^{(l)})
\]
Finally, the activation of the last layer produces the network's predicted
output:
\[
\hat{y} = a^{(L)}.
\]

The network is trained by backpropagating the error of a loss function 
$\mathbf{L}$, which quantifies how close the predicted output is to the 
desired output. Optimization algorithms such as gradient descent and 
backpropagation are then applied to update the weights of the connections 
$\mathbf{W}^{(l)}$ and the bias terms $\mathbf{b}^{(l)}$, with the aim 
of minimizing this loss and improving the model’s performance.

Denote the output of the DFN layer by
\emph{\textbf{F\textsubscript{final}}}\hspace{0pt}. The likelihood of
the brain tumor presence is represented by the sigmoid activation
function:
\begin{quote}
\begin{equation}
\mathbf{P} = \boldsymbol{\sigma} \left( \mathbf{F}_{\mathrm{final}} \right)
\end{equation}
\end{quote}

Where $\boldsymbol{\sigma}(x)$ is the sigmoid activation function as

follows:
\begin{quote}
\begin{equation}
\sigma(x) = \frac{1}{1 + e^{-x}}
\end{equation}
\end{quote}

The final decision can be made based on the probability output
\emph{\textbf{p}}.

\begin{quote}
\begin{equation}
y =
\begin{cases}
1, & \text{if } p > 0.5 \ \text{(tumor detected)} \\
0, & \text{if } p \leq 0.5 \ \text{(no tumor detected)}
\end{cases}
\end{equation}
\end{quote}

\subsection{Optimization and Training Setup}

The model was optimized using the Adam optimizer with a learning rate of \(1 \times 10^{-4}\), binary cross-entropy loss, and mini-batch size of 32. Early stopping and learning rate scheduling were employed to prevent overfitting. On average, training converged within 25–30 epochs

\section{Result and Discussion}
In the study, we applied the Resnet50 with Global Average Pooling and linear projection to extract features. Then we applied DFN to reduce the feature and classification. The models were evaluated with 2 datasets, where each dataset was assessed based on accuracy, precision, recall, and F1-score.

\subsection{Evaluation Metrics}
Evaluation metrics are used to evaluate the performance of the models compared to other models' performance. 

\begin{equation}
\text{Accuracy} = \frac{TP + TN}{TP + TN + FP + FN}
\end{equation}

\begin{equation}
\text{Precision} = \frac{TP}{TP + FP}
\end{equation}

\begin{equation}
\text{Recall} = \frac{TP}{TP + FN}
\end{equation}

\begin{equation}
F1-score = 2 \times \frac{\text{Precision} \times \text{Recall}}{\text{Precision} + \text{Recall}}
\end{equation}
\subsection{Ablation Study}
Beside DFN we also applied other combinations of deep learning to observe the classification accuracy rate.  1D CNN and FCNN are describe below as the classifier. 
\subsubsection{1D CNN}
The 1D CNN  body takes a feature vector resulting from GAP and conducts
convolution‐based operations to capture the spatial hierarchies of the
features \cite{ref28}\cite{ref31}\cite{ref32}. Because the input is 1D (i.e., vector
after GAP), the Convolution down-samples and captures local patterns.
1D Convolution over an input 1D tensor:
\begin{quote}
\begin{equation}
\mathbf{F}_{\mathrm{conv}}(t) =
\sum_{k=1}^{K} \mathbf{W}_{k} \cdot
\mathbf{F}_{\mathrm{avg}}(t + k) + b
\end{equation}
\end{quote}
Where \emph{\textbf{W\textsubscript{k}}} is the Convolution kernels,
\emph{\textbf{t}} is the timestep index, and \emph{\textbf{K}} is the
number of filters.

\subsubsection{FCNN}
FCNN is an ordinary feedforward neural network with a number of layers
in which all the neurons in the layer are connected to all the neurons
in the next layer \cite{ref49}. The FCNN processes these features in another
way and produces the final prediction. The forward function of a fully connected layer can be formulated as:
\begin{quote}
\begin{equation}
\mathbf{F}_{\mathrm{next}} = \mathbf{W} \cdot \mathbf{F}_{\mathrm{prev}} + b
\end{equation}
\end{quote}
Where \emph{\textbf{W}} is the weight matrix and \emph{\textbf{F\textsubscript{prev}}}\hspace{0pt},
\emph{\textbf{F\textsubscript{next}}\hspace{0pt}} are feature vectors in the previous layer and the output of the current layer, respectively.

\subsection{Performance of Newly Created MMCBT dataset:}

\begin{enumerate}
\def\labelenumi{\arabic{enumi}.}
\item
  \textbf{1D CNN }: The 1D CNN  model achieved an accuracy of
  \textbf{95.85\%,} where precision of \textbf{96.33\%,} recall of
  \textbf{95.15\%,} and an F1-score of \textbf{95.74\%.} While the model
  performed well, there is room for improvement in terms of recall.
\item
  \textbf{FCNN}: The Fully Connected Neural Network (FCNN) model
  outperformed 1D CNN  with an accuracy of \textbf{97.44\%,} precision of
  \textbf{96.03\%,} recall of \textbf{98.73\%,} and F1-score of
  \textbf{97.36\%.} The increase in recall shows the
  model\textquotesingle s stronger ability to identify positive
  instances.
\item
  \textbf{DFN}: The Deep Feedforward Network (DFN) classifier
  demonstrated the best performance with an accuracy of
  \textbf{97.57\%,} precision of \textbf{96.67\%,} recall of
  \textbf{98.42\%,} and an F1-score \textbf{o}f \textbf{97.54\%.} This classifier exhibited the highest accuracy and strong precision and recall values.
\end{enumerate}

\begin{table}[htbp]
\centering
\caption{\centering Performance of Newly Created MMCBT dataset}
\label{tab1}

\begin{tabular}{@{}lcccc@{}}
\toprule
\textbf{Model} & \textbf{Accuracy (\%)} & \textbf{Precision (\%)} & \textbf{Recall (\%)} & \textbf{F1-Score (\%)} \\
\midrule
1D CNN  & 95.85 & 96.33 & 95.15 & 95.74 \\
FCNN   & 97.44 & 96.03 & 98.73 & 97.36 \\
DFN    & 97.57 & 96.67 & 98.42 & 97.54 \\
\bottomrule
\end{tabular}
\end{table}

\begin{figure*}[htbp]
    \centering
    \begin{subfigure}{0.3\textwidth}
    \centering
    \includegraphics[width=\textwidth]{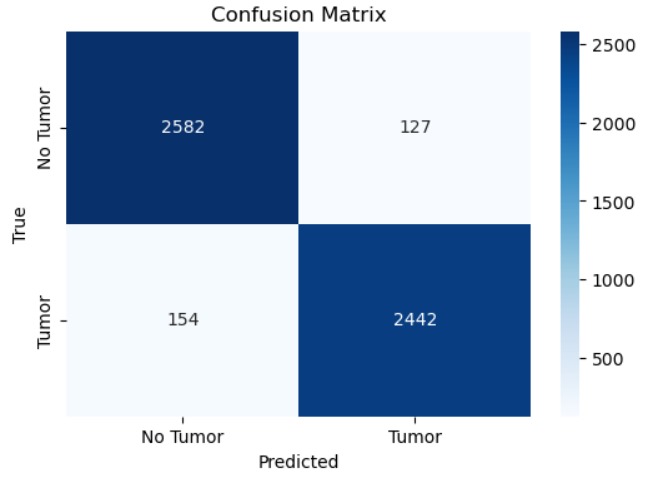}
    \subcaption[]{1D CNN }
    \end{subfigure}
    \begin{subfigure}{0.3\textwidth}
    \centering
    \includegraphics[width=\textwidth]{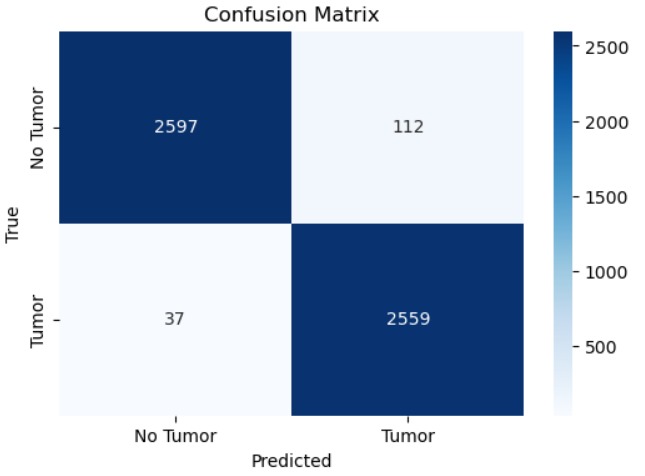}
    \subcaption[]{FCNN}
    \end{subfigure}
    \begin{subfigure}{0.3\textwidth}
    \centering
    \includegraphics[width=\textwidth]{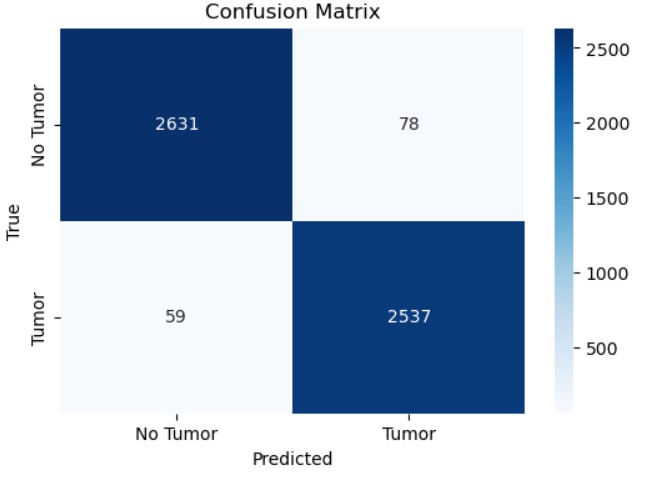}
    \subcaption[]{DFN}
    \end{subfigure}
    \caption{\centering Confusion Matrices of Classifiers using Newly Created MMCBT Dataset.}
    \label{confusionMatrix-1}
\end{figure*}

\begin{figure*}[htbp]
    \centering
    \begin{subfigure}{0.3\textwidth}
    \centering
    \includegraphics[width=\textwidth]{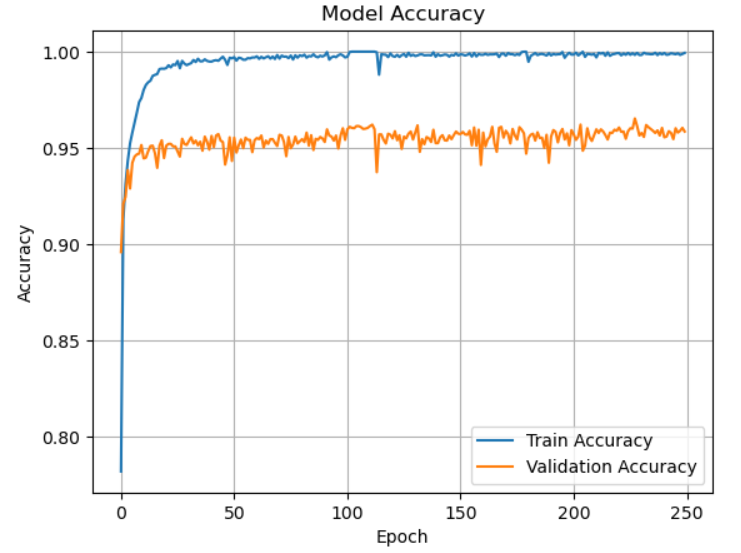}
    \subcaption[]{CNN}
    \end{subfigure}
     \begin{subfigure}{0.3\textwidth}
    \centering
    \includegraphics[width=\textwidth]{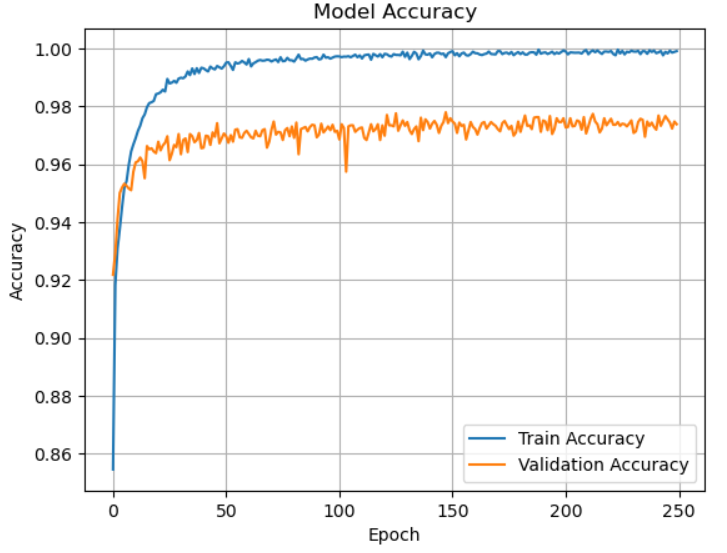}
    \subcaption[]{FCNN}
    \end{subfigure}
     \begin{subfigure}{0.3\textwidth}
    \centering
    \includegraphics[width=\textwidth]{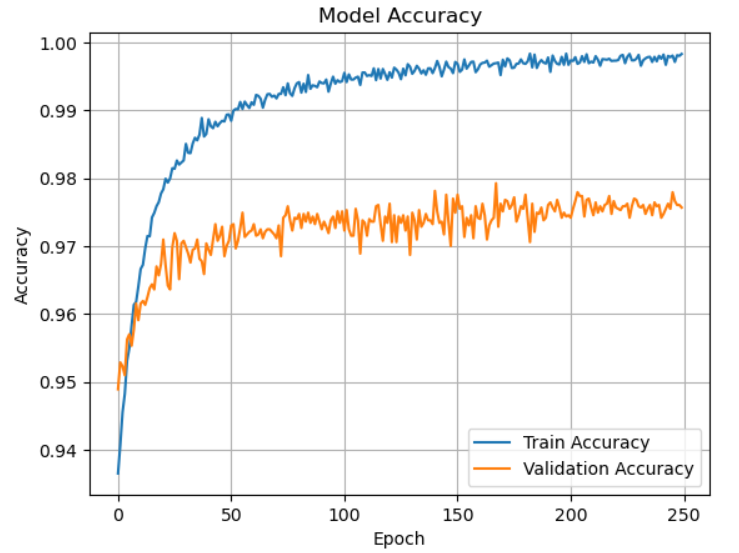}
    \subcaption[]{DFN}
    \end{subfigure}
    \caption{\centering Training and Validation Accuracy Plots using Newly Created MMCBT Dataset.}
    \label{accuracy-1}
\end{figure*}

\vspace{1em}

\begin{figure*}[htbp]
    \centering
    \begin{subfigure}{0.3\textwidth}
    \centering
    \includegraphics[width=\textwidth]{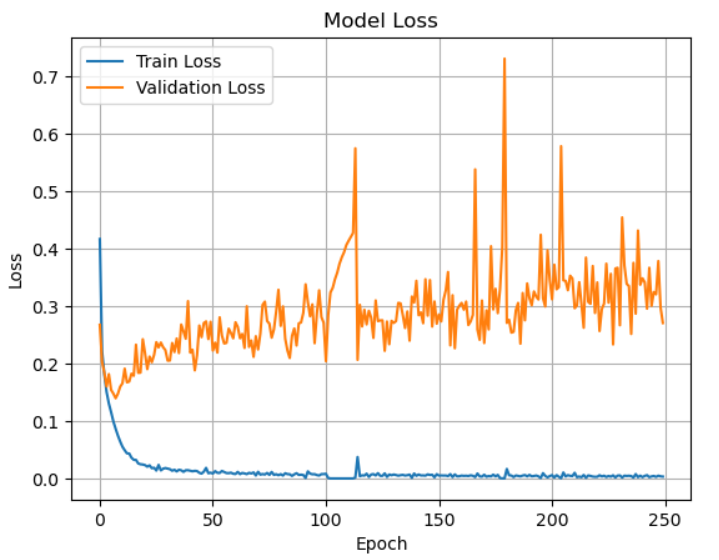}
    \subcaption[]{1D CNN }
    \end{subfigure}
    \begin{subfigure}{0.3\textwidth}
    \centering
    \includegraphics[width=\textwidth]{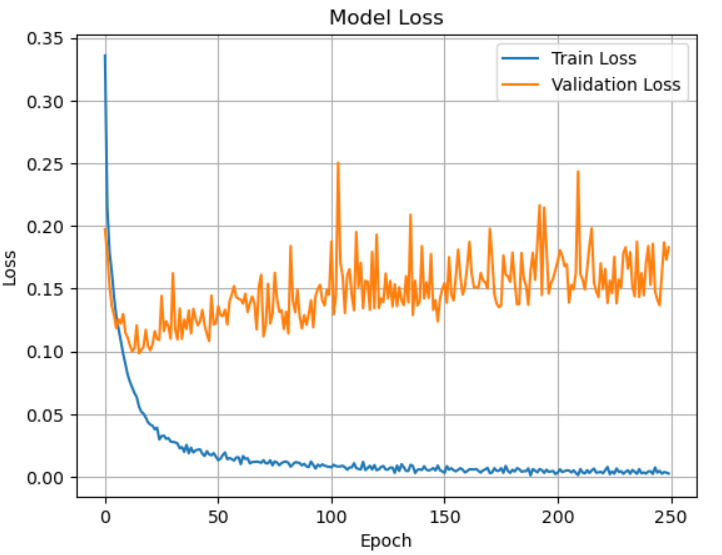}
    \subcaption[]{FCNN}
    \end{subfigure}
    \begin{subfigure}{0.3\textwidth}
    \centering
    \includegraphics[width=\textwidth]{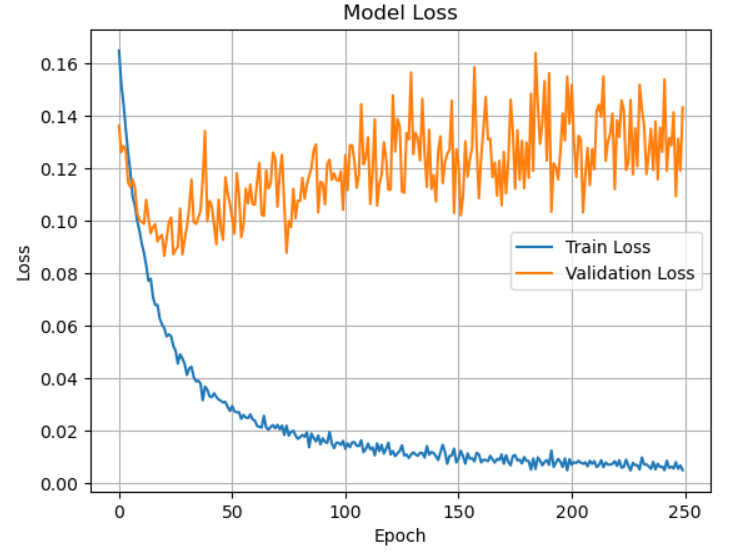}
    \subcaption[]{DFN}
    \end{subfigure}
    \caption{\centering Training and Validation Loss Plots using Newly Created MMCBT Dataset.}
    \label{loss-1}
    
\end{figure*}

Figure \ref{accuracy-1} and Figure \ref{loss-1} describe the training and validation accuracy
and loss. In 1D CNN  model, It achieved 99.97\% training accuracy and
95.85\% validation accuracy from where at the same time the training
loss 0.12\% is and validation loss is 27.06\%. In FCNN (Fully Connected
Neural Network) model achieved 99.89\% training accuracy and 97.44\%
validation accuracy, where at the same time the training loss 0.34\% is
and validation loss is 16.3\%. In DFN model also describe the training
and validation accuracy and the training and validation loss of CNN. It
achieved 99.82\% training accuracy and 97.57\% validation accuracy,
where at the same time the training loss 0.50\% is and validation loss is 14.34\%.\\

\renewcommand{\arraystretch}{1.5} 

\begin{table*}[htbp]
\centering
\caption{\centering Comparison of Performance in terms of Accuracy between Proposed Method and Existing Work on Brain Tumor Detection}
\label{tab2}
\begin{tabular}{@{}p{6cm}p{4cm}c@{}}
\toprule
\textbf{Author(s)} & \textbf{Classifier} & \textbf{Accuracy (\%)} \\
\midrule
Mahesh Swami et al.(2020)\cite{ref4} & k-means clustering & 87.50 \\
Mircea Gurbin et al.(2019)\cite{ref14} & SVM & 92.00 \\
Tomasila, G. et al.(2020)\cite{ref19} & RBF NN, DNN & 83.00 \\
Sakshi Ahuja et al.(2020)\cite{ref28} & CNN, AlexNet & 93.30 \\
Shraddha S. More et al.(2021)\cite{ref32} & NN, CNN & 89.00 \\
Manikandan Ramachandran et al. (2023)\cite{ref34} & SVM & 90.00 \\
Hein Tun Zaw et al. (2019)\cite{ref54} & Naïve Bayes & 94.00 \\
Khizar Abbas et al. (2019)\cite{ref45} & LIPC & 95.00 \\
Arashdeep Kaur (2016)\cite{ref43} & Fuzzy c-means & 90.57 \\
Sanjay Kumar C K et al.(2020)\cite{ref52} & SVM & 95.00 \\
\multirow{3}{*}{\textbf{Proposed}} & 1D CNN  & 95.85\% \\
& FCNN & 97.44\% \\
& DFN & 97.57\% \\
\bottomrule
\end{tabular}
\end{table*}

The table \ref{tab2}, presents several research efforts along with their
corresponding classification models and the accuracy they achieved. It
includes multiple studies conducted by various authors who applied
different machine learning methods to classification tasks. Each entry
lists the author names along with the classifiers used and the accuracy
results. The research incorporates a range of classifiers, including
k-means clustering, SVM, RBF Neural Networks (NN), Deep Neural Networks
(DNN), Convolutional Neural Networks (CNN), AlexNet, Naïve Bayes, LIPC,
fuzzy c-means, and other modern techniques such as 1D CNN, Fully
Convolutional Neural Networks (FCNN), and Deep Feature Networks (DFN).
The classifiers achieved accuracy rates between 83\% and 97.5\%, with
the highest performance recorded by the proposed models: 1D CNN at
95.85\% and DFN at 97.5\%

\begin{table}[htbp]
\centering
\caption{\centering Model Accuracy and Timing Metrics of Newly Created MMCBT Dataset}
\label{tab3}
\begin{tabular}{@{}lccc@{}}
\toprule
\textbf{Model} & \textbf{Accuracy (\%)} & \textbf{Time/Epoch (s)} & \textbf{Test Time/Image (ms)} \\
\midrule
VGG16         & 96.55 & 1541 & 377 \\
ResNet50      & 96.77 & 1623 & 700 \\
CNN-2D        & 97.15 & 1429 & 197 \\
DenseNet121   & 89.14 & 1390 & 1300 \\
CNN-1D        & 95.85 & 45   & 178 \\
FCNN          & 97.44 & 7    & 184 \\
DFN           & 97.57 & 8    & 193 \\
\bottomrule
\end{tabular}
\end{table}

The Table \ref{tab3} analysis of the dataset reveals that DFN performs the
best with the highest accuracy of 97.57\% and relatively fast processing
times, taking only 8 seconds per epoch and 193 ms per image. FCNN
follows closely with an accuracy of 97.44\%, being the fastest in terms
of training time at just 7 seconds per epoch, though its test time per
image is slightly slower at 184 ms. CNN-2D also achieves a high accuracy
of 97.15\% with a fast test time of 197 ms, but it takes 1429 seconds
per epoch, which is slower compared to FCNN and DFN. On the other hand,
ResNet50 and VGG16 show moderate accuracies of 96.77\% and 96.55\%,
respectively, but they are significantly slower, particularly in
testing, with ResNet50 taking 700 ms per image and VGG16 taking 377 ms.
DenseNet121 has the lowest accuracy of 89.14\% and also has the slowest
performance, with a test time of 1300 ms per image and a training time
of 1390 seconds per epoch, making it the least efficient among the
models.

\begin{figure}[htbp]
    \centering
    \includegraphics[width=0.5\textwidth]{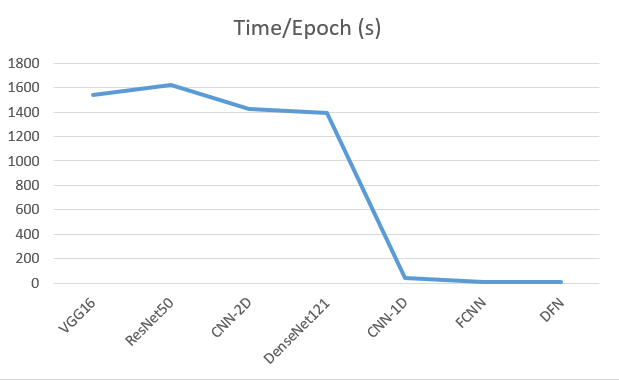}   
    \caption{Comparison of Training Time of Different Algorithms using Newly Created MMCBT Dataset.}
    \label{comparison-1}
\end{figure}
Figure \ref{comparison-1} shows that the time taken of training to complete each epoch shows a wide range across different models.It can be observed that CNNs such as ResNet-50 and VGG16 have the highest training times, 1623 and 1541 seconds, respectively, due to their computational complexity. CNN-2D and DenseNet-121 also consume considerable elapsed time, logging 1429 and 1390 seconds per epoch. The CNN-1D model is significantly faster, taking only 45 seconds. FCNN and DFN are the two fastest models to train, taking 7 and 8 seconds per epoch due to their light design and quick convergence. In general, this dataset yields longer training times for deeper models and shorter training for simpler ones.
\subsection{Performance of public BTM Dataset}
\begin{enumerate}
\def\labelenumi{\arabic{enumi}.}
\item
  \textbf{1D CNN }: The 1D CNN  model showed significant improvement on
  the public BTM Dataset, achieving an accuracy of \textbf{99.45\%,}
  precision of \textbf{99.19\%,} recall of \textbf{99.69\%,} and an
  F1-score of \textbf{99.44\%.} This model demonstrated exceptional
  recall performance, which is crucial for high-quality classification.
\item
  \textbf{FCNN}: The FCNN classifier reached an accuracy of
  \textbf{99.65\%,} precision of \textbf{99.80\%,} recall of
  \textbf{99.49\%,} and F1-score of \textbf{99.64\%.} This model
  excelled in precision, while maintaining a very high recall and
  F1-score.
\item
  \textbf{DFN}: The DFN classifier achieved the best performance across
  all metrics on the public BTM Dataset as shown in Table 2, with an
  accuracy of \textbf{99.80\%,} precision of \textbf{99.80\%,} recall of
  \textbf{99.80\%,} and an F1-score of \textbf{99.80\%.} The DFN model
  delivered perfect balance across all performance metrics, resulting in
  the highest overall classification performance.\\
\end{enumerate}

\begin{table}[htbp]
\centering
\caption{\centering Performance of public BTM Dataset}
\label{tab4}
\begin{tabular}{@{}lcccc@{}}
\toprule
\textbf{Model} & \textbf{Accuracy (\%)} & \textbf{Precision (\%)} & \textbf{Recall (\%)} & \textbf{F1-Score (\%)} \\
\midrule
1D CNN  & 99.45 & 99.19 & 99.69 & 99.44 \\
FCNN   & 99.65 & 99.80 & 99.49 & 99.64 \\
DFN    & 99.80 & 99.80 & 99.80 & 99.80 \\
\bottomrule
\end{tabular}
\end{table}

\begin{figure*}[htbp]
    \centering
    \begin{subfigure}{0.3\textwidth}
    \centering
    \includegraphics[width=\textwidth]{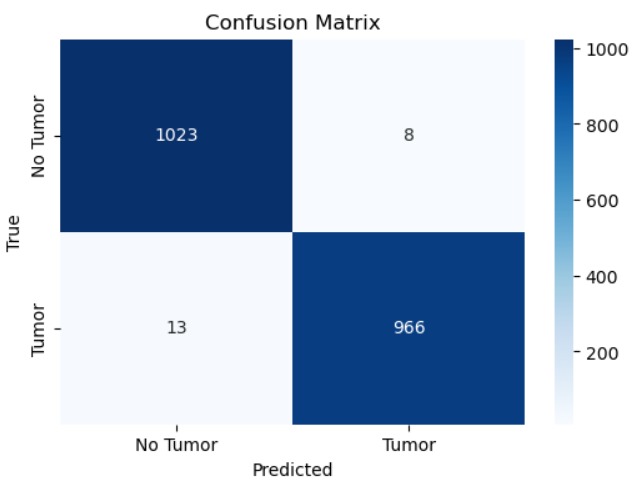}
    \subcaption[]{1D CNN }
    \end{subfigure}
    \begin{subfigure}{0.3\textwidth}
    \centering
    \includegraphics[width=\textwidth]{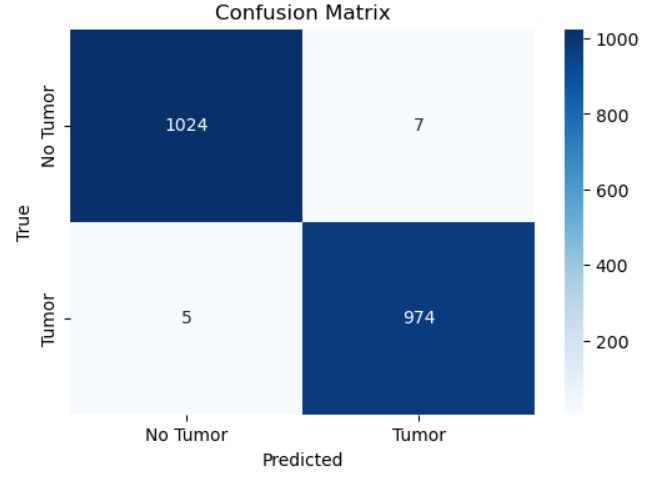}
    \subcaption[]{1D CNN }
    \end{subfigure}
    \begin{subfigure}{0.3\textwidth}
    \centering
    \includegraphics[width=\textwidth]{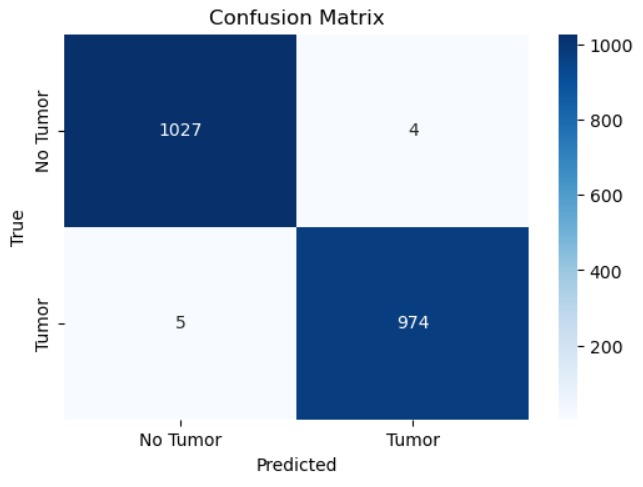}
    \subcaption[]{1D CNN }
    \end{subfigure}
    \caption{\centering Confusion Matrices of Classifiers using public BTM Dataset.}
    \label{confusionMatrix-2}
    
\end{figure*}

\begin{figure*}[htbp]
    \centering
    \begin{subfigure}{0.3\textwidth}
    \centering
    \includegraphics[width=\textwidth]{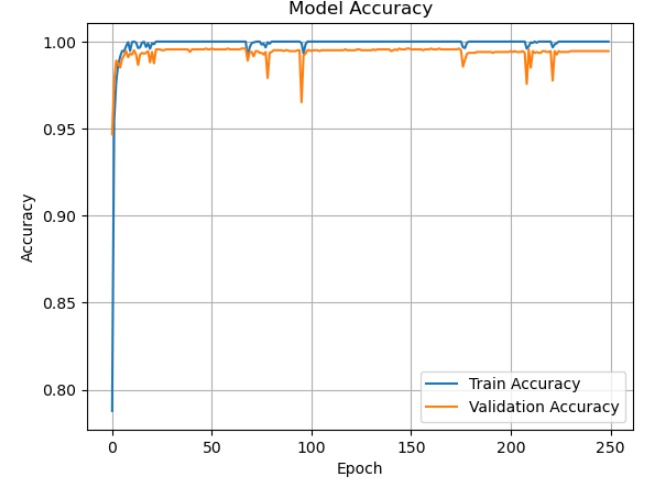}
    \subcaption[]{1D CNN }
    \end{subfigure}
    \begin{subfigure}{0.3\textwidth}
    \centering
    \includegraphics[width=\textwidth]{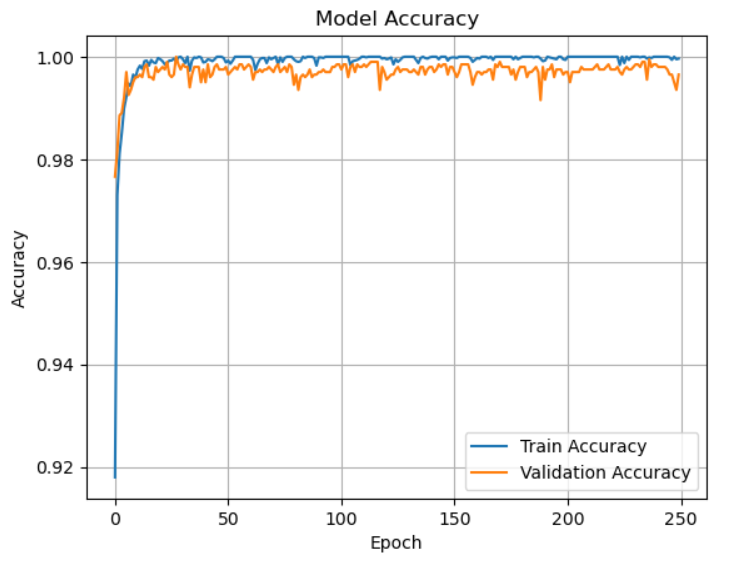}
    \subcaption[]{FCNN}
    \end{subfigure}
    \begin{subfigure}{0.3\textwidth}
    \centering
    \includegraphics[width=\textwidth]{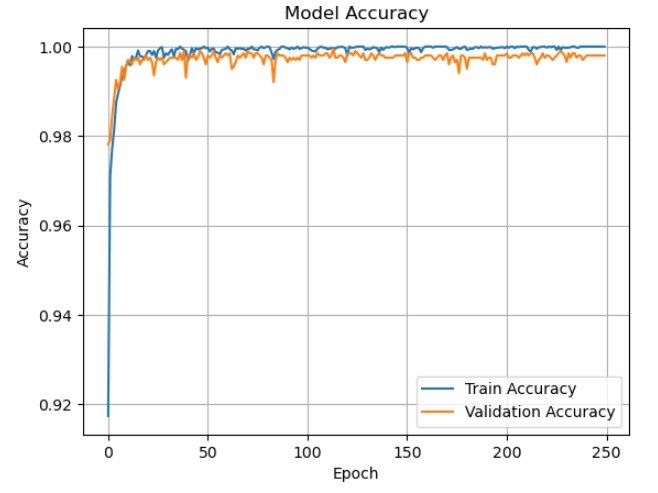}
    \subcaption[]{DFN}
    \end{subfigure}
    \caption{\centering Training and Validation Accuracy Plots using public BTM Dataset.}
\label{accuracy-2}
\end{figure*}

\begin{figure*}[htbp]
    \centering
    \begin{subfigure}{0.3\textwidth}
    \centering
    \includegraphics[width=\textwidth]{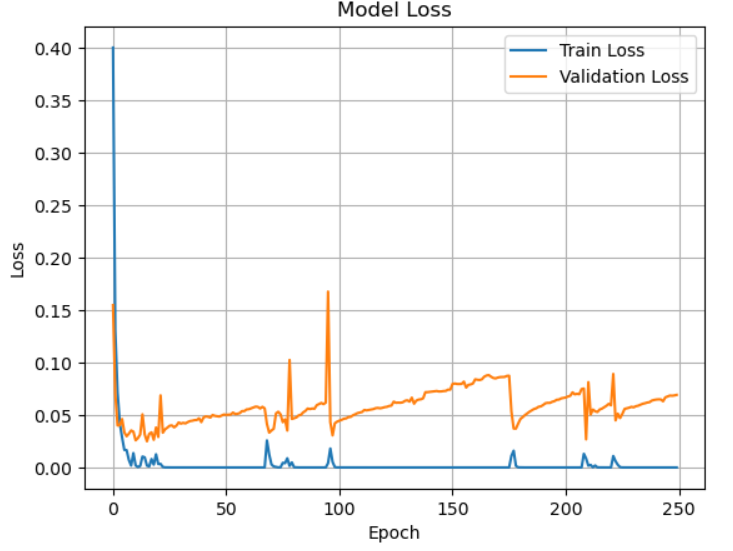}
    \subcaption[]{1D CNN }
    \end{subfigure}
    \begin{subfigure}{0.3\textwidth}
    \centering
    \includegraphics[width=\textwidth]{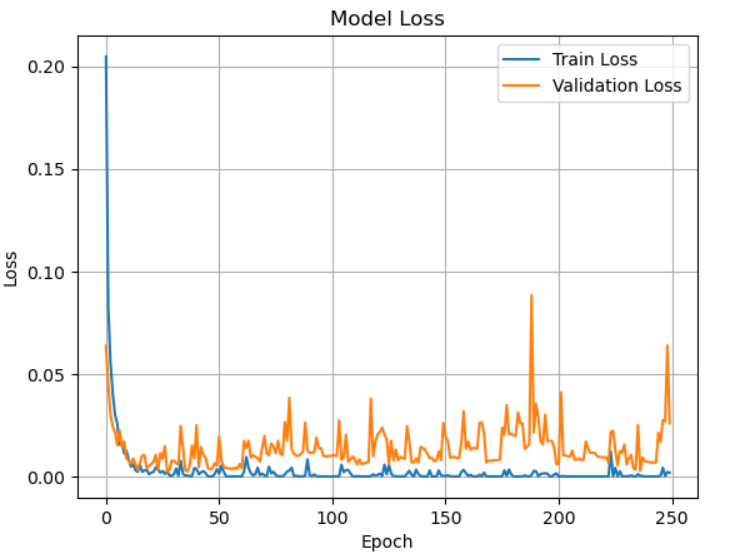}
    \subcaption[]{FCNN}
    \end{subfigure}
    \begin{subfigure}{0.3\textwidth}
    \centering
    \includegraphics[width=\textwidth]{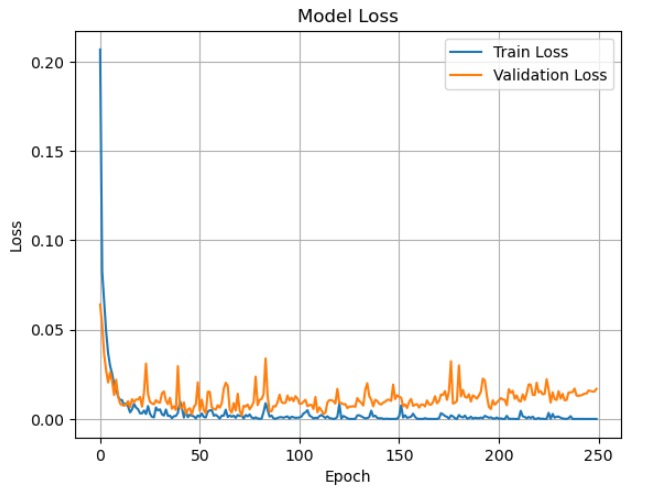}
    \subcaption[]{DFN}
    \end{subfigure}
    \caption{\centering Training and Validation Loss Plots 
using public BTM Dataset.}
\label{loss-2}
\end{figure*}

Figure \ref{accuracy-2} and Figure \ref{loss-2} describe the training and validation accuracy and loss. In CNN model, It achieved 100\% training accuracy and 99.45\% validation accuracy, where at the same time the training loss 8.47e-06\% is and validation loss is 6.92\%. In FCNN model, it achieved 99.93\% training accuracy and 99.65\% validation accuracy, where at the same
time the training loss 0. 26\% is and validation loss is 2.59\%. In DFN
model also describe the training and validation accuracy and loss of
CNN. It achieved 100\% training accuracy and 99.80\% validation
accuracy, where at the same time the training loss 0.0002\% is and
validation loss is 14.34\%.

\begin{table*}[htbp]
\centering
\caption{\centering Comparison of Performance in Terms of Accuracy between Proposed Method and Existing Work on public BTM Dataset}
\label{tab5}
\begin{tabular}{@{}p{4cm}p{4.5cm}p{3.5cm}c@{}}
\toprule
\textbf{Authors} & \textbf{Feature Extraction Method} & \textbf{Classifier} & \textbf{Accuracy (\%)} \\
\midrule
Pattabirama Mohan et al.(2024).\cite{ref63} & SWT and GLCM & MobileNet & 93.33 \\
Md. Zakir Hossain Zamil et al. (2025)\cite{ref60} & Convolutional Layer & CNN & 94.28 \\
Md. Ashraful Sharker Nirob et al. (2025)\cite{ref61} & NM & DenseNet201, InceptionV3, MobileNetV3 & 87.21 / 82.30 / 92.01 \\
Sayed Masuma Ali et al. (2025)\cite{ref62} & FFV & Random Forest & 97.00 \\
\multirow{3}{*}{\textbf{Proposed}} & \multirow{3}{*}{Pre-trained CNN, GAP, Linear Projection} & CNN & 99.45\% \\
& & FCNN & 99.65\% \\
& & DFN & 99.80\% \\
\bottomrule
\end{tabular}
\end{table*}

The Table \ref{tab5} presents several research efforts along with their
corresponding classification models and the accuracy they achieved. It
includes multiple studies conducted by various authors who applied
different machine learning methods to classification tasks. Each entry
lists the author names along with the classifiers used and the accuracy
results. The research incorporates a range of classifiers, including
k-means clustering, SVM, RBF Neural Networks (NN), Deep Neural Networks
(DNN), Convolutional Neural Networks (CNN), AlexNet, Naïve Bayes, LIPC,
fuzzy c-means, and other modern techniques such as 1D CNN, Fully
Convolutional Neural Networks (FCNN), and Deep Feature Networks (DFN).
The classifiers achieved accuracy rates between 83\% and 97.5\%, with
the highest performance recorded by the proposed models: 1D CNN at
95.85\% and DFN at 97.5\%

\begin{table}[htbp]
\centering
\caption{\centering Model Accuracy and Timing Metrics of public BTM Dataset}
\label{tab6}
\begin{tabular}{@{}lccc@{}}
\toprule
\textbf{Model} & \textbf{Accuracy (\%)} & \textbf{Time/Epoch (s)} & \textbf{Test Time/Image (ms)} \\
\midrule
VGG16       & 99.72 & 517  & 326  \\
ResNet50    & 99.69 & 652  & 689  \\
CNN-2D      & 99.08 & 494  & 193  \\
DenseNet121 & 97.75 & 483  & 1135 \\
CNN-1D      & 99.45 & 15   & 169  \\
FCNN        & 99.65 & 3    & 172  \\
DFN         & 99.80 & 3    & 189  \\
\bottomrule
\end{tabular}
\end{table}

\begin{figure}[htbp]
    \centering
    \includegraphics[width=0.5\textwidth]{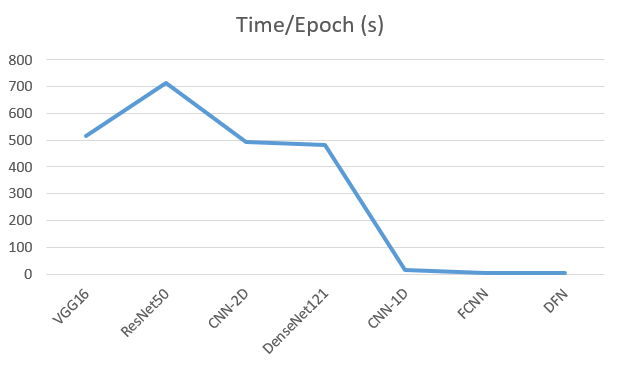}
    
    \caption{\centering Comparison of Training Time of Different Algorithms using public BTM Dataset.}
    \label{comparison-2}
\end{figure}

Figure \ref{comparison-2} shows the training time per epoch is much less. For deep models, the ResNet-50 model has the longest training time, at 712 seconds, followed by the VGG16 model of 517 seconds. CNN-2D and DenseNet121 also exhibit similarly medium-length training times, at 494 seconds and 483 seconds, respectively. Compared to the CNN-2D model, the CNN-1D model is much faster, with each epoch requiring only 15 seconds. The quickest results are achieved with FCNN and DFN, needing only 3 seconds per epoch.

In Table \ref{tab6} (public BTM Dataset), DFN stands out with the highest accuracy of
99.80\%, along with extremely fast training (3 seconds per epoch) and
test times (189 ms per image), making it the most efficient and balanced
model. VGG16 follows closely with 99.72\% accuracy, though it takes
significantly longer for training (517 seconds per epoch) and testing
(326 ms per image). CNN-2D and CNN-1D also perform well, with accuracies
of 99.08\% and 99.45\%, respectively, but CNN-1D is notably faster in
training (15 seconds per epoch) and testing (169 ms per image). FCNN
offers 99.65\% accuracy, with exceptionally fast training (3 seconds per
epoch) and a test time of 172 ms per image, making it a strong contender
for efficiency. DenseNet121, with the lowest accuracy of 97.75\%, also
has the slowest test time (1135 ms per image), making it less efficient
than the other models.

The results indicate that the DFN classifier achieved the highest
performance on both datasets. On the Newly Created Newly Created MMCBT Dataset, DFN achieved an
accuracy of \textbf{97.57\%,} while on the public BTM Dataset, it reached an
accuracy of \textbf{99.80\%,} with a perfectly balanced performance
across all metrics. FCNN also performed exceptionally well, particularly
in terms of recall, while 1D CNN  demonstrated solid results but with
slightly lower performance in comparison.

These findings demonstrate that the DFN classifier, particularly when
combined with feature extraction techniques like ResNet50, GAP, and
Linear Projection, offers robust and consistent performance across
different datasets.

\section{Conclusion}
In this paper, we proposed an automatic and efficient deep-learning framework for brain tumor detection from MRI scans. The approach leverages a pre-trained ResNet50 model for feature extraction, enhanced by Global Average Pooling (GAP) and linear projection to generate compact and discriminative image representations. A novel Dense-Dropout sequence was introduced to improve non-linear feature learning, reduce overfitting, and enhance model robustness. A key contribution of this work is the development of the Mymensingh Medical College Brain Tumor (MMCBT) dataset, which provides a clinically verified and balanced resource comprising 209 subjects and over 16,000 MRI images. This dataset addresses the scarcity of reliable brain tumor MRI data and establishes a strong foundation for future research in automated diagnosis. Extensive experiments conducted on both the MMCBT dataset and the publicly available BTM dataset demonstrated the effectiveness of the proposed framework, achieving accuracies of 97.57\% and 99.80\%, respectively, along with strong precision, recall, and F1-scores. These results underscore the potential of our approach as a robust and reliable tool for clinical decision support. The proposed framework combines methodological innovation with valuable dataset creation, significantly advancing research on brain tumor detection. While some limitations remain, this work contributes a step forward toward the development of automated diagnostic systems capable of assisting radiologists, improving diagnostic efficiency, and ultimately supporting better patient outcomes.

\bibliographystyle{unsrt}
\bibliography{Reference}

\begin{thebibliography}{10}

\bibitem{ref41}
Ghazanfar Latif, M.~Mohsin Butt, Adil~H. Khan, Omair Butt, and D.~N. F.~Awang Iskandar.
\newblock Multiclass brain glioma tumor classification using block-based 3d wavelet features of mr images.
\newblock In {\em 2017 4th International Conference on Electrical and Electronic Engineering (ICEEE)}, 2017.

\bibitem{ref18}
Sheeba Khan.
\newblock Detection of brain tumor using digital image processing.
\newblock {\em Journal of Health \& Medical Informatics}, 2022.

\bibitem{ref43}
Arashdeep Kaur.
\newblock An automatic brain tumor extraction system using different segmentation methods.
\newblock In {\em 2016 Second International Conference on Computational Intelligence \& Communication Technology (CICT)}, 2016.

\bibitem{ref45}
Khizar Abbas, Prince~Waqas Khan, Khan~Talha Ahmed, and Wang-Cheoul Song.
\newblock Automatic brain tumor detection in medical imaging using machine learning.
\newblock In {\em 2019 International Conference on Information and Communication Technology Convergence (ICTC)}, 2019.

\bibitem{ref46}
Sahar Ghanavati, Junning Li, Ting Liu, Paul~S. Babyn, Wendy Doda, and George Lampropoulos.
\newblock Automatic brain tumor detection in magnetic resonance images.
\newblock In {\em 2012 9th IEEE International Symposium on Biomedical Imaging (ISBI)}, 2012.

\bibitem{ref7}
K.~S. Thara and K.~Jasmine.
\newblock Brain tumor detection in mri images using pnn and grnn.
\newblock In {\em 2016 International Conference on Wireless Communications, Signal Processing and Networking (WiSPNET)}, 2016.

\bibitem{ref9}
Shobana G and Ranjith Balakrishnan.
\newblock Brain tumor diagnosis from mri feature analysis -- a comparative study.
\newblock In {\em 2015 International Conference on Innovations in Information, Embedded and Communication Systems (ICIIECS)}, 2015.

\bibitem{hassan2024residual_miah_alzh}
Najmul Hassan, Abu~Saleh Musa~Miah, and Jungpil Shin.
\newblock Residual-based multi-stage deep learning framework for computer-aided alzheimer’s disease detection.
\newblock {\em Journal of Imaging}, 10(6):141, 2024.

\bibitem{hassan2025neurological}
Najmul Hassan, Abu Saleh~Musa Miah, Yuichi Okuyama, and Jungpil Shin.
\newblock Neurological disorder recognition via comprehensive feature fusion by integrating deep learning and texture analysis.
\newblock {\em IEEE Open Journal of the Computer Society}, 2025.

\bibitem{hassan2025stacked_alzh_miah}
Najmul Hassan, Abu Saleh~Musa Miah, Kota Suzuki, Yuichi Okuyama, and Jungpil Shin.
\newblock Stacked cnn-based multichannel attention networks for alzheimer disease detection.
\newblock {\em Scientific Reports}, 15(1):5815, 2025.

\bibitem{Hassan_alzh_gradual_miah_}
Najmul Hassan, Abu Saleh~Musa Miah, Taro Suzuki, and Jungpil Shin.
\newblock Gradual variation-based dual-stream deep learning for spatial feature enhancement with dimensionality reduction in early alzheimer’s disease detection.
\newblock {\em IEEE Access}, 13:31701--31717, 2025.

\bibitem{haque2024multi_heart_disease}
Mahfuzul Haque, Abu Saleh~Musa Miah, Debashish Gupta, Md~Maruf Al~Hossain Prince, Tanzina Alam, Nusrat Sharmin, Mohammed~Sowket Ali, and Jungpil Shin.
\newblock Multi-class heart disease detection, classification, and prediction using machine learning models.
\newblock {\em arXiv preprint arXiv:2412.04792}, 2024.

\bibitem{ref38}
Amer Al-Badarneh, Hassan Najadat, and Ali~M. Alraziqi.
\newblock A classifier to detect tumor disease in mri brain images.
\newblock In {\em 2012 IEEE/ACM International Conference on Advances in Social Networks Analysis and Mining}, 2012.

\bibitem{miah2025methodologica_pd}
Abu Saleh~Musa Miah, Taro Suzuki, and Jungpil Shin.
\newblock A methodological and structural review of parkinson’s disease detection across diverse data modalities.
\newblock {\em IEEE Access}, 2025.

\bibitem{shin2025autism_miah}
Jungpil Shin, Abu Saleh~Musa Miah, Manato Kakizaki, Najmul Hassan, and Yoichi Tomioka.
\newblock Autism spectrum disorder detection using skeleton-based body movement analysis via dual-stream deep learning.
\newblock {\em Electronics}, 14(11):2231, 2025.

\bibitem{shin2025_pd_musa}
Jungpil Shin, Abu Saleh~Musa Miah, Koki Hirooka, Md~Al~Mehedi Hasan, and Md~Maniruzzaman.
\newblock Parkinson disease detection based on in-air dynamics feature extraction and selection using machine learning.
\newblock {\em Scientific Reports}, 15(1):28027, 2025.

\bibitem{matsumoto2025machine_musa_pd}
Masahiro Matsumoto, Abu Saleh~Musa Miah, Nobuyoshi Asai, and Jungpil Shin.
\newblock Machine learning-based differential diagnosis of parkinson’s disease using kinematic feature extraction and selection.
\newblock {\em IEEE Access}, 2025.

\bibitem{ref35}
Subhashis Banerjee, Sushmita Mitra, and B.~Uma Shankar.
\newblock Synergetic neuro-fuzzy feature selection and classification of brain tumors.
\newblock In {\em 2017 IEEE International Conference on Fuzzy Systems (FUZZ-IEEE)}, 2017.

\bibitem{ref42}
Neelum Noreen, Sellappan Palaniappan, Abdul Qayyum, Iftikhar Ahmad, Muhammad Imran, and Muhammad Shoaib.
\newblock A deep learning model based on concatenation approach for the diagnosis of brain tumor.
\newblock {\em IEEE Access}, 8, 2020.

\bibitem{ref44}
S.~Gayathri and D.~C. Joy~Winnie Wise.
\newblock Analyzing, detecting and automatic classification of different stages of brain tumor using region segmentation and support vector machine.
\newblock In {\em 2020 International Conference on Electronics and Sustainable Communication Systems (ICESC)}, 2020.

\bibitem{ref47}
K.~N. Deeksha, Deeksha M, Anagha~V. Girish, Anusha~S. Bhat, and Lakshmi H.
\newblock Classification of brain tumor and its types using convolutional neural network.
\newblock In {\em 2020 IEEE International Conference for Innovation in Technology (INOCON)}, 2020.

\bibitem{ref48}
Mahdis~Roshanfekr Rad and Alireza Sahab.
\newblock Diffusion magnetic resonance imaging for brain tumor detection with segmentation active contour.
\newblock In {\em 2017 IEEE International Conference on Cybernetics and Computational Intelligence (CyberneticsCom)}, 2017.

\bibitem{ref49}
Adekanmi~A. Adegun and Serestina Viriri.
\newblock Fcn-based densenet framework for automated detection and classification of skin lesions in dermoscopy images.
\newblock {\em IEEE Access}, 8, 2020.

\bibitem{ref50}
V.~P. Gladis~Pushpa Rathi and S.~Palani.
\newblock Detection and characterization of brain tumor using segmentation based on hsom, wavelet packet feature spaces and ann.
\newblock In {\em 2011 3rd International Conference on Electronics Computer Technology}, 2011.

\bibitem{ref51}
Khaleda~Akhter Sathi and Md.~Saiful Islam.
\newblock Hybrid feature extraction based brain tumor classification using an artificial neural network.
\newblock In {\em 2020 IEEE 5th International Conference on Computing Communication and Automation (ICCCA)}, 2020.

\bibitem{ref52}
Sanjay Kumar~C K and H.~D. Phaneendra.
\newblock Categorization of brain tumors using svm with hybridized local-global features.
\newblock In {\em 2020 Fourth International Conference on Computing Methodologies and Communication (ICCMC)}, 2020.

\bibitem{ref53}
Muhammad Nazir, Muhammad~Attique Khan, Tanzila Saba, and Amjad Rehman.
\newblock Brain tumor detection from mri images using multi-level wavelets.
\newblock In {\em 2019 International Conference on Computer and Information Sciences (ICCIS)}, 2019.

\bibitem{ref54}
Hein~Tun Zaw, Noppadol Maneerat, and Khin~Yadanar Win.
\newblock Brain tumor detection based on naïve bayes classification.
\newblock In {\em 2019 5th International Conference on Engineering, Applied Sciences and Technology (ICEAST)}, 2019.

\bibitem{ref55}
M.~Monica Subashini and Indra~Gandhi V.
\newblock An efficient non-invasive method for brain tumor grade analysis on mr images.
\newblock In {\em TENCON 2017 - 2017 IEEE Region 10 Conference}, 2017.

\bibitem{ref56}
Kang~Han Oh, Soo~Hyung Kim, and Myungeun Lee.
\newblock Tumor detection on brain mr images using regional features: Method and preliminary results.
\newblock In {\em 2015 21st Korea-Japan Joint Workshop on Frontiers of Computer Vision (FCV)}, 2015.

\bibitem{saeedi2023mri}
Soheila Saeedi, Sorayya Rezayi, Hamidreza Keshavarz, and Sharareh R.~Niakan~Kalhori.
\newblock Mri-based brain tumor detection using convolutional deep learning methods and chosen machine learning techniques.
\newblock {\em BMC Medical Informatics and Decision Making}, 23(1):16, 2023.

\bibitem{asiri2024optimized}
Abdullah~A Asiri, Toufique~Ahmed Soomro, Ahmed~Ali Shah, Ganna Pogrebna, Muhammad Irfan, and Saeed Alqahtani.
\newblock Optimized brain tumor detection: a dual-module approach for mri image enhancement and tumor classification.
\newblock {\em IEEE access}, 12:42868--42887, 2024.

\bibitem{vasavi2025hybrid}
G~Vasavi, Vaddadi~Vasudha Rani, Sreenu Ponnada, and S~Jyothi.
\newblock A hybrid efficientnet-dbnealexnet for brain tumor detection using mri images.
\newblock {\em Computational Biology and Chemistry}, 115:108279, 2025.

\bibitem{ref58}
Ayesha Ghaffar.
\newblock Brain tumor data, 2024.

\bibitem{ref28}
Sakshi Ahuja, B.~K. Panigrahi, and Tapan Gandhi.
\newblock Transfer learning based brain tumor detection and segmentation using superpixel technique.
\newblock In {\em 2020 International Conference on Contemporary Computing and Applications (IC3A)}, 2020.

\bibitem{ref31}
M.~Li, L.~Kuang, S.~Xu, and S.~Zhan{-}Guo.
\newblock Brain tumor detection based on multimodal information fusion and convolutional neural network.
\newblock {\em IEEE Access}, 7:180134--180146, 2019.

\bibitem{ref32}
Shraddha~S. More, Mansi~Ashok Mange, Mitheel~Sandip Sankhe, and Shwethali~Santosh Sahu.
\newblock Convolutional neural network based brain tumor detection.
\newblock In {\em Proceedings of the Fifth International Conference on Intelligent Computing and Control Systems (ICICCS 2021)}, 2021.

\bibitem{ref4}
Mahesh Swami and Divya Verma.
\newblock An algorithmic detection of brain tumour using image filtering and segmentation of various radiographs.
\newblock In {\em 2020 7th International Conference on Signal Processing and Integrated Networks (SPIN)}, 2020.

\bibitem{ref14}
Mircea Gurbin, Mihaela Lascu, and Dan Lascu.
\newblock Tumor detection and classification of mri brain image using different wavelet transforms and support vector machines.
\newblock In {\em 2019 42nd International Conference on Telecommunication and Signal Processing (TSP)}, 2019.

\bibitem{ref19}
G.~Tomasila and A.~W.~R. Emanuel.
\newblock Mri image processing method on brain tumors: A review.
\newblock In {\em Nucleation and Atmospheric Aerosols}, 2020.

\bibitem{ref34}
Manikandan Ramachandran, Rizwan Patan, Ambeshwar Kumar, Soheil Hosseini, and Amir~H. Gandomi.
\newblock Mutual informative mapreduce and minimum quadrangle classification for brain tumor big data.
\newblock {\em IEEE Transactions on Engineering Management}, 2023.

\bibitem{ref63}
Pattabirama Mohan and G.~Ramkumar.
\newblock Detection and classification of brain tumor using fine tuned mobile net algorithm.
\newblock In {\em 2024 3rd International Conference for Advancement in Technology (ICONAT)}, 2024.

\bibitem{ref60}
Md~Zakir~Hossain Zamil et~al.
\newblock Optimizing magnetic resonance imaging analysis for brain tumors: A lightweight neural network approach.
\newblock In {\em Proceedings of the ACM}, 2025.

\bibitem{ref61}
Md. Ashraful~Sharker Nirob et~al.
\newblock Attention-based multi-scale fusion for brain tumor classification with explainable ai.
\newblock In {\em 2025 International Conference on Electrical, Computer and Communication Engineering (ECCE)}, 2025.

\bibitem{ref62}
Sayed~Masuma Ali et~al.
\newblock Deep transfer learning framework for brain tumor detection in mri slices: A study.
\newblock In {\em 2025 International Conference on Frontier Technologies and Solutions (ICFTS)}, 2025.

\end{thebibliography}

\newcommand{\EOD}{}

\end{document}